\documentclass{article}

 \usepackage[preprint]{neurips_2026}

\usepackage[utf8]{inputenc} 
\usepackage[T1]{fontenc}    
\usepackage{hyperref}       
\usepackage{url}            
\usepackage{booktabs}       
\usepackage{amsfonts}       
\usepackage{nicefrac}       
\usepackage{microtype}      
\usepackage{xcolor}         

\usepackage{graphicx}
\usepackage{subcaption}
\usepackage{multirow}
\usepackage[table]{xcolor}
\usepackage{amsmath}
\usepackage{array}
\usepackage{enumitem}
\usepackage[normalem]{ulem}
\usepackage{algorithm}
\usepackage{algpseudocode}
\usepackage{wrapfig}

\title{GeoSVG-RL: Geometry-Aware Reinforcement Learning for Layout-Constrained Text-to-SVG Diagram Generation}

\author{%
Sifan~Li\textsuperscript{\dag}
\hspace{1em}
Yujun~Cai\textsuperscript{\ddag}
\hspace{1em}
Hongkai~Chen\textsuperscript{\S}\thanks{Corresponding author.}
\hspace{1em}
Yiwei~Wang\textsuperscript{\dag}
\\
\textsuperscript{\dag}University of California,~Merced
\hspace{1em}
\textsuperscript{\ddag}The University of Queensland
\\
\textsuperscript{\S}vivo Mobile Communication Co., Ltd.
\\
\texttt{sflijohn@foxmail.com, allenhkchen@gmail.com}
\\
\textcolor{magenta}{\texttt{https://github.com/johnnyZeppelin/GeoSVG-RL}}
}

\newcommand{\method}{GeoSVG-RL}

\begin{document}

\maketitle

\begin{abstract}
Generating structured, editable diagrams remains a significant challenge for contemporary large language models, despite their proficiency in general-purpose vector code generation. The primary difficulty lies in the structural fragility of the output; minor errors such as misaligned connector endpoints, text labels overlapping borders, or complex layouts drifting beyond the canvas boundaries render the resulting SVG files functionally unusable for professional applications. To address these issues, we introduce \method, a specialized reinforcement learning framework designed for layout-constrained text-to-SVG generation. Unlike standard training objectives that rely solely on maximizing token-level likelihood, our approach optimizes the policy against explicit, executable geometric feedback. The model first produces a structured layout plan that serves as a geometric contract for the subsequent generation of the SVG code. This code is then rendered through a browser-backed verifier, enabling the calculation of fine-grained rewards across six critical dimensions: rendering validity, canvas fitting, precise anchor placement, text containment, graph consistency, and code cleanliness. We utilize Group Relative Policy Optimization (GRPO) to refine the model, sampling multiple candidates per prompt to facilitate updates based on relative quality. Starting from a supervised warm-start phase on synthetic data, \method\ achieves substantial gains in structural reliability, particularly in arrow-anchor accuracy and text-in-box rates. Quantitative evaluations demonstrate that our method consistently outperforms current state-of-the-art systems in local geometric precision and the preservation of graph connectivity, providing a robust pathway toward automated yet reliable technical illustration.
\end{abstract}

\section{Introduction}
Scalable Vector Graphics (SVG) are an ideal format for technical diagrams, offering inherent scalability, ease of post-generation editing, and a transparent underlying code structure. Although recent Large Language Models (LLMs) and Vision-Language Models (VLMs) have shown significant progress in generating SVG from textual and visual prompts \citep{rodriguez2023starvector,xing2024llm4svg,yang2025omnisvg,wang2025svgen}, synthesizing structured box-arrow-text diagrams remains more difficult than creating icons or free-form illustrations. The quality of these diagrams relies on strict geometric constraints that are obvious to human readers but frequently violated by current models. Specifically, connectors must align with valid anchor points, labels must stay within boxes with proper padding, related modules must remain aligned, and the entire figure must fit within the canvas boundaries.

\begin{figure}
\centering
\includegraphics[width=\textwidth]{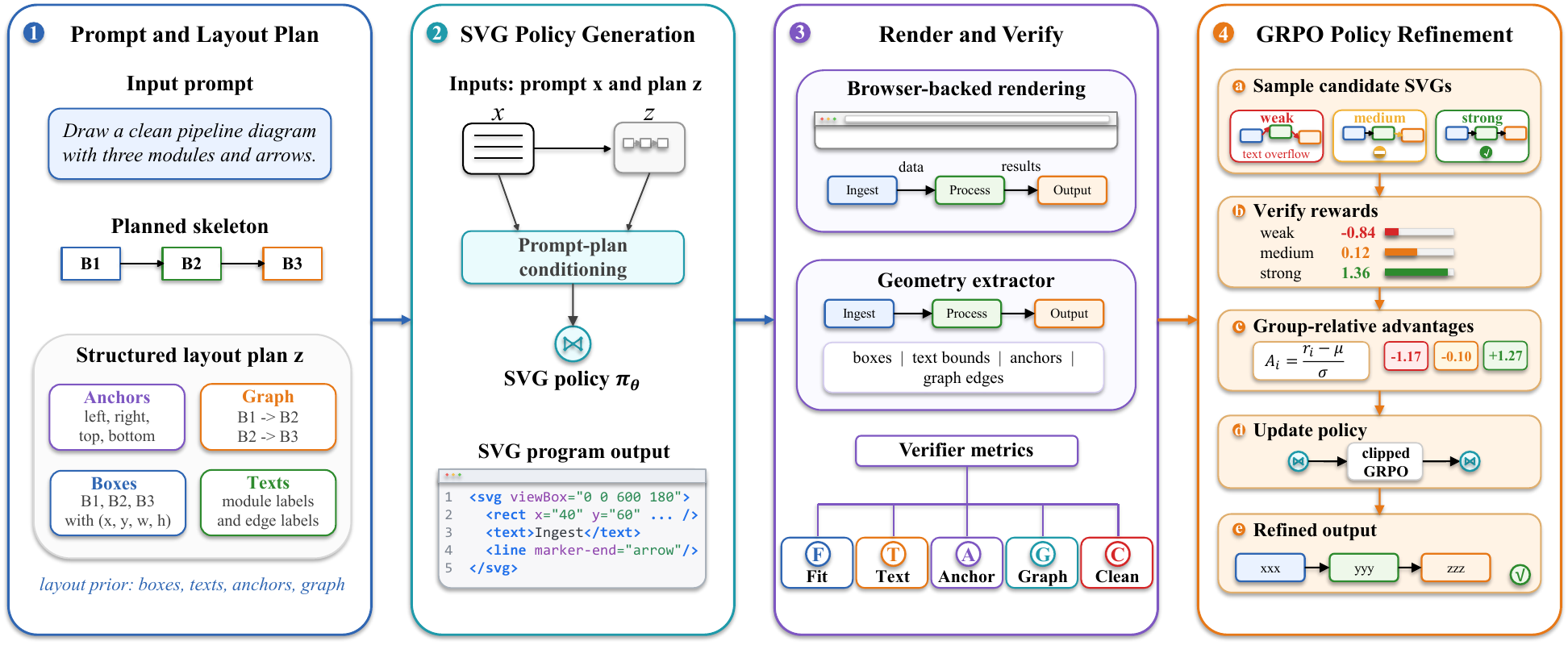}
\caption{The \method\ framework adopts a plan-then-generate approach, where a structured layout plan $z$ is first predicted from the textual prompt $x$. An SVG policy $\pi_\theta$ then generates SVG code conditioned on both $x$ and $z$. This code is rendered by a browser-backed verifier to extract bounding boxes, text boundaries, anchor points, and graph edges, which serve as the basis for computing geometry-aware rewards across dimensions such as canvas fit, text containment, anchor alignment, graph consistency, and code cleanliness. During training, the model samples multiple candidates for each prompt, converts verifier scores into group-relative advantages, and updates the policy via clipped GRPO. This method ensures the generation of diagrams that are structurally reliable and faithful to the intended layout.}
\vspace{-2em}
\label{fig:teaser}
\end{figure}

Geometric precision is where existing systems often fail. In scientific and technical illustrations, even small inaccuracies can hinder readability, complicate manual editing, and reduce the perceived reliability of the content \citep{zhu2026autofigure,lin2026autofigureedit,he2026vfig}. The core challenge is not that modern generators cannot produce plausible SVG code, but that standard training objectives do not directly encourage the structural properties necessary for diagrammatic integrity. While next-token prediction treats SVG generation as a sequence modeling task, broad image-level metrics are typically too coarse to capture whether a connector is accurately anchored or whether a label has sufficient clearance from its container.

To address these limitations, we treat the generation of diagrammatic SVG as a constrained program synthesis problem. Since SVG is executable code, the output can be rendered, parsed, and evaluated using explicit spatial heuristics. This approach allows us to train the model based on desired behavioral outcomes rather than hoping these properties emerge indirectly from sequence modeling.

Our method, \method, implements this concept through a structured pipeline. The model predicts an explicit layout plan before generating the corresponding SVG code. A browser-based verifier then renders the SVG, extracts the geometry, and evaluates the output according to execution validity, canvas fitting, anchor placement, text containment, graph consistency, and code cleanliness. After an initial warm start on synthetic plan-SVG pairs, the model is refined via GRPO \citep{shao2024deepseekmath} using rewards from the verifier.


The main contributions of this work are summarized as follows:
\begin{itemize}
    \item \textbf{Problem Reformulation}: We formulate the generation of layout-constrained text-to-SVG diagrams as a constrained program synthesis problem, enabling the integration of executable geometric feedback into the training loop.
    \item \textbf{Framework Development}: We introduce \method, a structured pipeline encompassing planning, generation, and browser-backed verification. Unlike traditional sequence modeling, \method\ optimizes rendered geometry rather than raw token likelihood.
    \item \textbf{Technical Methodology}: We provide a practical implementation of verifier-backed reinforcement learning for SVG generation, detailing synthetic data generation, robust structural parsing, and optimization via Group Relative Policy Optimization (GRPO).
    \item \textbf{Empirical Evaluation}: We present a geometry-aware evaluation suite and demonstrate that \method\ significantly outperforms supervised baselines and existing state-of-the-art models in local geometric precision and graph connectivity.
\end{itemize}

\section{Related Work}
\paragraph{SVG Generation via LLMs and VLMs.}
Scalable Vector Graphics (SVG) are increasingly treated as a task of structured sequence generation rather than a simple target for post-processing. StarVector pioneered this area by introducing a multimodal model for both image-to-SVG and text-to-SVG synthesis \citep{rodriguez2023starvector}. LLM4SVG subsequently improved SVG understanding and generation using semantic tokenization and instruction tuning \citep{xing2024llm4svg}. UniSVG expanded the field by providing a unified dataset of 525k items, enabling Large Multimodal Models (LMMs) to bridge the gap between SVG comprehension and production \citep{10.1145/3746027.3758269}. OmniSVG scaled this approach using the MMSVG-2M dataset \citep{yang2025omnisvg}, while SVGen examined SVG-specific supervision, chain-of-thought annotation, and reinforcement learning \citep{wang2025svgen}. Other works, including CTRL-S, IntroSVG, and VectorGym, have explored multi-task learning, rendering-aware optimization, and benchmark design \citep{wang2026ctrls,wang2026introsvg,rodriguez2026vectorgym}.

Recent methods have begun to incorporate reasoning and rendering feedback into the training process. Reason-SVG introduced a ``Drawing-with-Thought'' paradigm, where the model generates design rationales before producing code to activate reasoning patterns necessary for structural validity \citep{xing2026reasonsvg}. Similarly, RLRF (Reinforcement Learning from Rendering Feedback) addresses the non-differentiable nature of rendering by using a VLM-based reinforcement learning framework to optimize visual fidelity and semantic coherence \citep{rodriguez2026renderingaware}. While these systems provide strong baselines, they generally prioritize semantic alignment and aesthetics over the strict, layout-dependent geometric constraints, such as precise anchor-point alignment, required for professional technical diagrams.

\paragraph{Scientific Figure and Diagram Generation.}
Scientific illustration requires high structural fidelity. AutoFigure and AutoFigure-Edit generate publication-quality figures from long-form scientific text, focusing on editable SVG outputs \citep{zhu2026autofigure,lin2026autofigureedit}. VFig investigates converting complex technical figures into SVG and provides a dataset and evaluation suite for this task \citep{he2026vfig}. Beyond generation, understanding the semantic structure of these diagrams is still a challenge; TechING addresses this with a large-scale synthetic corpus for training VLMs to interpret technical illustrations like flowcharts \citep{nadeem-etal-2026-teching}. This work is closely related but focuses specifically on layout-constrained diagrams and the training signals needed to ensure geometry is both visually plausible and structurally accurate for editing.

\paragraph{Evaluation of Structured Diagrams.}
Standard image metrics are often insufficient for evaluating diagrams. DiagramEval suggests that generated diagrams should be evaluated as graphs, using node and path alignment rather than just image similarity \citep{liang2025diagrameval}. SVG benchmarks like SVGenius and VectorGym take a similar view, advocating for structure-aware evaluation \citep{chen2025svgenius,rodriguez2026vectorgym,zhu2026structural}. Furthermore, SVGauge emphasizes alignment with human cognition \citep{zini2025svgauge}. For editing, SVGEditBench established a framework to assess how LLMs modify SVG code for tasks like color or shape transformation \citep{Nishina_2024_CVPR}, which was later expanded by SVGEditBench V2 to include complex, instruction-based tasks \citep{nishina2025svg}. Our evaluation suite follows this direction but emphasizes executable geometric feedback at the rendering level, specifically measuring connector precision and text container integrity within constrained layouts.

\begin{figure}
\centering
\includegraphics[width=\textwidth]{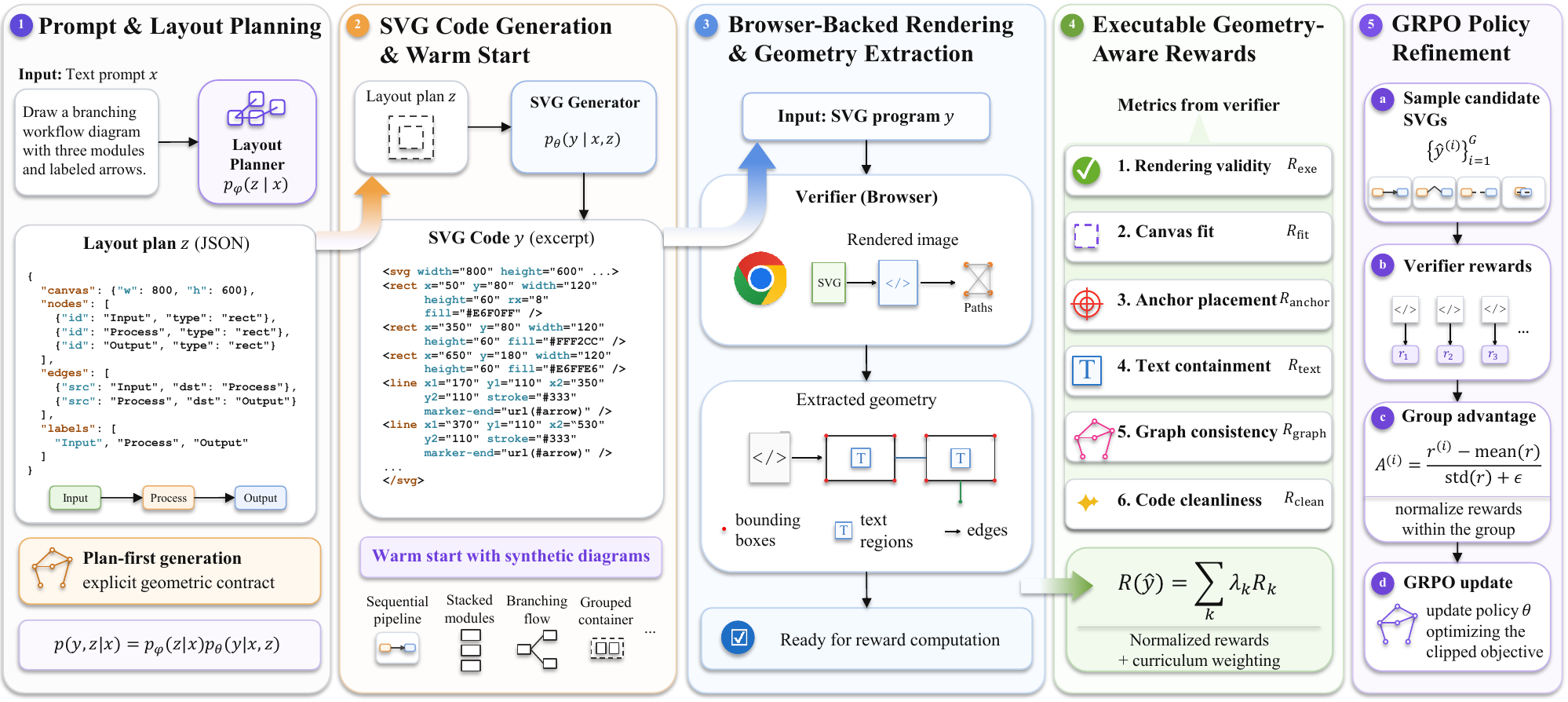}
\caption{
Overview of the \method\ framework. Given a textual prompt $x$, the model first predicts a structured layout plan $z$, which then guides the generation of executable SVG code $y$. The initial capacity for SVG generation is developed through a supervised warm-start phase using synthetic diagram-layout-SVG tuples. During the reinforcement learning stage, a browser-based verifier renders each sampled SVG to extract precise geometric features, including bounding boxes, text regions, connector anchors, and graph edges. These signals form the basis for computing geometry-aware rewards that account for rendering validity, canvas fit, anchor placement, text containment, graph consistency, and code cleanliness. Within each prompt group, multiple candidates are evaluated to determine group-relative advantages, which are subsequently used to refine the SVG policy via clipped GRPO.
}
\label{fig:pipeline}
\end{figure}

\section{Method}
Given a textual prompt $x$ describing a structured diagram, we aim to generate SVG code $y$ that renders a valid diagram within a target canvas $C$. Each output consists of typed graphical elements, such as bounding boxes, text nodes, and connectors. Unlike free-form vector illustrations, these diagrams must satisfy explicit spatial constraints: every connector must terminate at a valid anchor point, each text node must be contained within its assigned region with sufficient padding, and all visible elements must remain within the boundaries of $C$.

Let $r(y)$ denote the rasterized rendering of $y$ and $a(y)$ represent the parsed SVG abstract syntax tree (AST). During training, we use a latent layout specification $z$ for synthetic supervision, which details element types, target bounding boxes, valid anchors, and adjacency relations. The learning objective is to maximize both the conditional probability $p(y \mid x)$ and the expected geometry-aware reward $R(x, y, z)$ derived from post-rendering analysis.


\subsection{Layout-Plan-First Generation}
We factorize the conditional distribution as:
\begin{equation}
    p(y,z \mid x) = p_{\phi}(z \mid x)\, p_{\theta}(y \mid x,z),
\end{equation}
where $p_{\phi}$ represents a layout planner and $p_{\theta}$ denotes an SVG code generator. The plan $z$ encompasses the canvas dimensions, node bounding boxes, text assignments, connector endpoints, and graph connectivity. A representative node entry is defined as:
\begin{equation}
    n_i = (\text{id}_i, \text{type}_i, x_i, y_i, w_i, h_i, s_i),
\end{equation}
where $s_i$ denotes the textual label. A connector entry specifies the identifiers for the source and destination nodes along with anchor types, such as left-center or bottom-center.

This plan-first factorization mitigates coordinate drift and provides the generator with an explicit geometric contract. Furthermore, this approach simplifies downstream verification, as the generated SVG can be evaluated against a structured target rather than solely a reference token sequence.

\subsection{Supervised Warm Start}


The policy is initialized from a pretrained autoregressive code model rather than from scratch. This base model already understands structured formats like XML and HTML. We perform a warm start using a synthetic dataset of diagram prompts, layout plans, ground-truth SVG programs, and geometric labels.

The synthetic data are generated procedurally by sampling various diagram families, such as sequential pipelines, stacked modules, and branching flows. For each graph template, a layout engine assigns coordinates, text regions, and connector routes. The system produces a natural-language prompt, a structured plan, and a ground-truth SVG, alongside corrupted variants featuring box shifts or misalignments. These samples help the model learn SVG syntax and the mapping from text to editable programs. The supervised objective for a triplet $(x, z^*, y^*)$ is:
\begin{equation}
\mathcal{L}_{\text{SFT}} = -\sum_{t=1}^{T}\log p_{\theta}(y^*_t \mid y^*_{<t}, x, z^*).
\end{equation}

\subsection{Browser-Backed Rendering and Geometry Extraction}

\paragraph{Rendering Path.}
SVG code is rendered through a browser engine to ensure high fidelity. In practice, a headless browser, such as Chromium, provides the most reliable processing of text layouts, font metrics, and complex transformations. This is particularly critical for text-intensive diagrams, as the actual dimensions of a text region depend on font rendering rather than raw coordinate attributes alone.

\paragraph{Structure Path.}
The SVG XML is parsed into a normalized geometric representation. The parser identifies primitives such as \texttt{rect}, \texttt{text}, and \texttt{path}, resolves transformations into a global coordinate system, and converts elements into canonical objects. This enables the recovery of graph structures and the assessment of code cleanliness directly from the vector program. This bifurcated approach is intentional: text containment and canvas overflow are assessed post-rendering, while graph connectivity and code quality are evaluated via the parsed structure.

\paragraph{Rewards.}
For a generated SVG $\hat y$, the verifier calculates a weighted sum of executable rewards:
\begin{equation}
R(\hat y) = \sum_i \lambda_i R_i.
\end{equation}
To enhance execution validity, we design a binary gate to ensure that malformed SVG outputs do not receive positive rewards:
\begin{equation}
R_{\text{exec}} = \mathbb{I}[\hat y \text{ is valid and renderable}].
\end{equation}
For canvas fitting, let $B_{\text{all}}$ denote the union of bounding boxes for all visible elements and $C$ represent the target canvas. We define a fit reward and an overflow penalty as follows:
\begin{equation}
R_{\text{fit}} = \mathbb{I}[B_{\text{all}} \subseteq C],
\qquad
R_{\text{overflow}} = -\frac{\mathrm{Area}(B_{\text{all}} \setminus C)}{\mathrm{Area}(B_{\text{all}})+\epsilon}.
\end{equation}
Regarding arrow anchoring, for each predicted endpoint $\hat a_k$ and reference anchor $a_k$, we compute the accuracy and the associated error:
\begin{equation}
R_{\text{anchor-acc}} = \frac{1}{M}\sum_{k=1}^{M}\mathbb{I}[\|\hat a_k-a_k\|_2 \leq \tau],
\qquad
R_{\text{anchor-err}} = -\frac{1}{M}\sum_{k=1}^{M}\frac{\|\hat a_k-a_k\|_2}{d_k}.
\end{equation}
For text containment and padding, let $\hat T_j$ be the rendered text box for instance $j$ and $B_j$ its assigned container. We define:
\begin{equation}
R_{\text{text-in-box}} = \frac{1}{K}\sum_{j=1}^{K}\mathbb{I}[\hat T_j \subseteq B_j].
\end{equation}
If $m_j$ represents the minimum distance between the text box and the container boundary, and $p$ denotes the required padding, then:
\begin{equation}
R_{\text{padding}} = -\frac{1}{K}\sum_{j=1}^{K}\mathbb{I}[m_j < p].
\end{equation}
To optimize graph connectivity, let $\hat{\mathcal{E}}$ be the predicted edge set and $\mathcal{E}$ represent the ground-truth set. We compute the F1 score:
\begin{equation}
R_{\text{graph}} = \frac{2PR}{P+R+\epsilon}.
\end{equation}
Beyond the finalized figure, the quality of the code itself is significant. We define SVG Cleanliness to incentivize the use of semantically meaningful primitives while penalizing fragmented code:
\begin{equation}
R_{\text{clean}} = \frac{\#\{\texttt{rect},\texttt{text},\texttt{line},\ldots\}}{\#\{\text{all elements}\}+\epsilon}.
\end{equation}

\subsection{Reward Normalization and Weighting}
Each reward component is normalized to $[0,1]$ or $[-1,0]$. We use a curriculum-based weighting strategy: training initially prioritizes local geometric correctness, such as anchoring, before increasing the emphasis on global canvas fitting.

\subsection{GRPO for SVG Policy Refinement}
Following the supervised warm start, the generator is refined using Group Relative Policy Optimization (GRPO). For a given prompt $x$, a layout plan $\hat z \sim p_{\phi}(\cdot\mid x)$ is sampled, followed by a group of SVG candidates $\{\hat y^{(i)}\}_{i=1}^{G} \sim p_{\theta}(\cdot\mid x,\hat z)$. Group-relative advantages $A^{(i)}$ are computed by normalizing the rewards within the group:
\begin{equation}
A^{(i)} = \frac{r^{(i)} - \mathrm{mean}(r)}{\mathrm{std}(r)+\epsilon}.
\end{equation}
We optimize a clipped objective:
\begin{equation}
\mathcal{L}_{\text{GRPO}} = \mathbb{E}\Big[ \min\big(\rho_i A^{(i)}, \quad \mathrm{clip}(\rho_i,1-\eta,1+\eta)A^{(i)}\big) \Big].
\end{equation}


The training procedure is outlined in Algorithm~\ref{alg:training}. We use a headless browser for rendering and a standard XML parser for extraction. During reinforcement learning, we found that starting with small group sizes and low-temperature sampling improves stability.

\begin{algorithm}[t]
\caption{Training \method}
\label{alg:training}
\begin{algorithmic}[1]
\Require Training set $\mathcal{D}=\{(x,z^*,y^*)\}$, planner $p_{\phi}$, generator $p_{\theta}$, verifier $\mathcal{V}$
\For{supervised training iterations}
    \State Sample $(x,z^*,y^*) \sim \mathcal{D}$
    \State Update $p_{\phi}$ with $\log p_{\phi}(z^*\mid x)$
    \State Update $p_{\theta}$ with $\log p_{\theta}(y^*\mid x,z^*)$
\EndFor
\For{GRPO refinement iterations}
    \State Sample prompt $x$
    \State Predict layout plan $\hat z \sim p_{\phi}(\cdot\mid x)$
    \State Sample a group of SVG candidates $\{\hat y^{(i)}\}_{i=1}^{G}$
    \For{$i=1$ to $G$}
        \State Obtain reward $r^{(i)} = \mathcal{V}(\hat y^{(i)})$
    \EndFor
    \State Compute group-relative advantages $A^{(i)}$
    \State Update $\theta$ with clipped GRPO objective
\EndFor
\end{algorithmic}
\end{algorithm}

\section{Experiments}
\subsection{Evaluation Metrics}
To quantitatively assess the performance of \method, we establish a comprehensive evaluation suite comprising ten geometry-aware metrics.
These metrics are categorized into execution validity, global layout integrity, and local geometric precision.

\paragraph{Execution and Global Fit.} 
The fundamental requirement for a generated diagram is its renderability. We define the \textbf{Render Success Rate (RSR)} as the proportion of valid SVG programs within the total number of test samples $N$:
\begin{equation}
\mathrm{RSR} = \frac{1}{N} \sum_{i=1}^{N} \mathbb{I}[y_i \text{ is a valid SVG program}].
\end{equation}
To evaluate whether the generated content remains within the intended workspace, we define the \textbf{Global Fit Rate (GFR)} and \textbf{Overflow Area Ratio (OAR)}:
\begin{equation}
\mathrm{GFR} = \mathbb{I}[B_{\text{all}} \subseteq C], \quad \mathrm{OAR} = \frac{\mathrm{Area}(B_{\text{all}} \setminus C)}{\mathrm{Area}(B_{\text{all}}) + \epsilon},
\end{equation}
where $B_{\text{all}}$ represents the union bounding box of all rendered elements and $C$ denotes the target canvas. Additionally, the \textbf{Element-In-Canvas Rate (EICR)} measures the proportion of individual elements $B_j$ that are fully contained within $C$:
\begin{equation}
\mathrm{EICR} = \frac{1}{|\mathcal{B}|} \sum_{B_j \in \mathcal{B}} \mathbb{I}[B_j \subseteq C].
\end{equation}

\paragraph{Local Geometric Precision.}
The structural utility of a diagram depends upon the precise alignment of connectors and text. We evaluate \textbf{Arrow Anchor Accuracy (AAcc)} and \textbf{Anchor Endpoint Error (AEE)} as follows:
\begin{equation}
\mathrm{AAcc} = \frac{1}{M} \sum_{k=1}^{M} \mathbb{I}[\|\hat{a}_k - a_k\|_2 \le \tau], \quad \mathrm{AEE} = \frac{1}{M} \sum_{k=1}^{M} \frac{\|\hat{a}_k - a_k\|_2}{d_k},
\end{equation}
where $\hat{a}_k$ and $a_k$ denote the predicted and ground-truth anchor coordinates, respectively, and $d_k$ serves as a normalization factor. For text integrity, the \textbf{Text-In-Box Rate (TBR)} and \textbf{Text Padding Violation Rate (TPVR)} are defined as:
\begin{equation}
\mathrm{TBR} = \frac{1}{K} \sum_{j=1}^{K} \mathbb{I}[\hat{T}_j \subseteq B_j], \quad \mathrm{TPVR} = \frac{1}{K} \sum_{j=1}^{K} \mathbb{I}[\mathrm{dist}(\hat{T}_j, \partial B_j) < p],
\end{equation}
where $\hat{T}_j$ is the rendered text bounding box, $B_j$ is its assigned container, and $p$ represents the minimum required padding.

\paragraph{Structural and Code Quality.} 
To ensure the diagram correctly represents the intended information flow, we compute the \textbf{Edge Connectivity F1 (E-F1)} score:
\begin{equation}
\mathrm{E\mbox{-}F1} = \frac{2 \cdot P \cdot R}{P + R + \epsilon},
\end{equation}
where precision $P$ and recall $R$ are determined by comparing the predicted edge set $\hat{\mathcal{E}}$ with the ground-truth set $\mathcal{E}$. Finally, \textbf{SVG Cleanliness (Clean)} quantifies the ratio of semantically meaningful primitives to the total number of geometric elements.

\subsection{Data and Split Protocol}
Our experiments use a fixed split of a procedurally generated diagram corpus. The training set is used for supervised warm start and GRPO refinement, the validation set for hyperparameter tuning and early stopping, and the test set exclusively for the final evaluation in Table~\ref{tab:svg_geometry_main}. To prevent data leakage, we define the split based on random seeds, graph templates, and prompt variations, ensuring the test set contains novel combinations of node labels, branching structures, and layout configurations.

The benchmark focuses on box-arrow-text diagrams, which highlight the geometric failures that motivate this study while remaining suitable for automatic verification. These include horizontal pipelines, stacked modules, grouped containers, and branching process flows. For each sample, we maintain the prompt, layout plan, reference SVG, rendered image, and geometric metadata such as bounding boxes, anchors, and edge lists.


\subsection{Baselines and Evaluation Protocol}
We compare \method\ against \textbf{VFig} \citep{he2026vfig}, a vectorization-oriented system, and \textbf{AutoFigure-Edit} \citep{lin2026autofigureedit}, which focuses on scientific illustrations. To ensure a fair comparison, all models generate a single SVG per prompt using a shared wrapper and fixed decoding settings, with self-consistency disabled. Any non-SVG explanatory text is removed to isolate the code artifact for analysis.

For the baselines, we design an evaluation protocol. The evaluation involves normalizing all outputs to SVG and processing them through our browser-based rendering and extraction pipeline. For systems that generate intermediate raster images, we only assess the final exported SVG code. This protocol, as detailed in Table~\ref{tab:svg_geometry_main}, applies uniform geometry-aware metrics to all models. This approach accounts for the inherent differences in their original design objectives, such as vectorization versus illustration, by focusing exclusively on final structural reliability.

\begin{table}[t]
  \caption{
Performance comparison across ten geometry-aware SVG generation metrics. 
Higher values indicate better performance except for OAR, AEE, and TPVR, where lower values are preferred. 
The best results are shown in \textbf{bold}, and the second-best results are \underline{underlined}.
}
  \label{tab:svg_geometry_main}
  \centering
  \setlength{\tabcolsep}{4.2pt}
  \small
  \resizebox{\textwidth}{!}{%
  \begin{tabular}{lcccccccccc}
    \toprule
    Model
    & RSR$\uparrow$
    & GFR$\uparrow$
    & OAR$\downarrow$
    & EICR$\uparrow$
    & AAcc$\uparrow$
    & AEE$\downarrow$
    & TBR$\uparrow$
    & TPVR$\downarrow$
    & E-F1$\uparrow$
    & Clean$\uparrow$ \\
    \midrule
    VFig
    & 90.2
    & 88.4
    & 3.5
    & 91.2
    & 76.6
    & 0.124
    & 81.8
    & 9.6
    & 85.7
    & 89.5 \\

    AutoFigure-Edit
    & \cellcolor{green!8}\underline{91.5}
    & \cellcolor{green!8}\underline{89.3}
    & \cellcolor{green!25}\textbf{2.2}
    & \cellcolor{green!8}\underline{92.7}
    & \cellcolor{green!8}\underline{77.4}
    & \cellcolor{green!8}\underline{0.105}
    & \cellcolor{green!8}\underline{81.6}
    & \cellcolor{green!8}\underline{6.1}
    & \cellcolor{green!8}\underline{86.9}
    & \cellcolor{green!8}\underline{90.2} \\

    \method~(Ours)
    & \cellcolor{green!25}\textbf{92.3}
    & \cellcolor{green!25}\textbf{89.7}
    & \cellcolor{green!8}\underline{2.9}
    & \cellcolor{green!25}\textbf{93.4}
    & \cellcolor{green!25}\textbf{78.6}
    & \cellcolor{green!25}\textbf{0.096}
    & \cellcolor{green!25}\textbf{83.0}
    & \cellcolor{green!25}\textbf{3.3}
    & \cellcolor{green!25}\textbf{90.4}
    & \cellcolor{green!25}\textbf{91.8} \\
    \bottomrule
  \end{tabular}%
  }
  \vspace{-1em}
\end{table}

\subsection{Main Results}
Table~\ref{tab:svg_geometry_main} compares the performance of various methods across ten geometry-aware metrics. \method\ achieves the strongest overall results, ranking first on nine metrics and second on OAR. Compared to VFig and AutoFigure-Edit, our approach yields significant improvements in both local geometric accuracy and structural quality.

\begin{wrapfigure}{r}{0.5\textwidth} 
    \centering
    \vspace{-1em}
    \includegraphics[width=\linewidth]{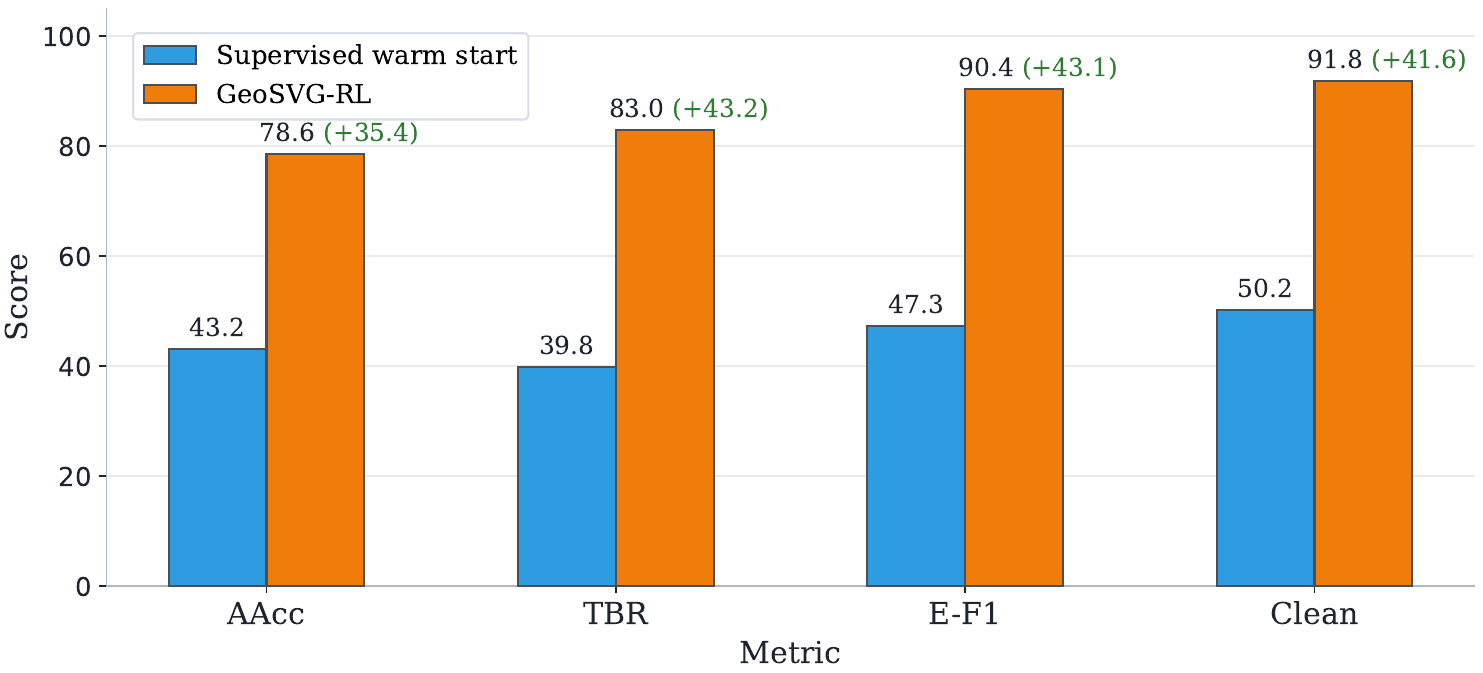}
    \caption{Performance gains of \method\ over the supervised warm start.}
    \vspace{-2em}
    \label{fig:bars}
\end{wrapfigure}

Specifically, our method \method\ achieves \textbf{78.6} AAcc, \textbf{0.096} AEE, \textbf{83.0} TBR, \textbf{3.3} TPVR, and \textbf{90.4} E-F1, demonstrating substantial gains in arrow anchoring, text containment, and graph consistency. These results suggest that executable rewards effectively optimize the constraints essential for diagram usability. Beyond local alignment, \method\ also leads in RSR (\textbf{92.3}), GFR (\textbf{89.7}), EICR (\textbf{93.4}), and Clean (\textbf{91.8}), which indicates that the reinforcement learning process improves both rendering validity and code quality.

The only exception is OAR, where \method\ (\underline{2.9}) follows AutoFigure-Edit (\textbf{2.2}). While reducing overflow remains an objective for future work, \method\ still outperforms VFig and maintains top scores in GFR and EICR. This remaining gap likely reflects the specific difficulty of minimizing residual overflow in dense layouts rather than a general failure in global control.

\subsection{Analysis}
Our evaluation shows that geometry-aware RL provides broad benefits across nearly all metrics. The most notable improvements occur in constraints that are difficult to capture through token-level supervision, such as anchor placement and graph connectivity. Compared to the supervised warm-start phase, \method\ increases AAcc from 43.2 to 78.6, TBR from 39.8 to 83.0, and E-F1 from 47.3 to 90.4. These gains validate the effectiveness of verifier feedback in optimizing geometric properties that are only indirectly represented in the SVG code.

Furthermore, the leading performance of \method\ in RSR, GFR, EICR, and TPVR proves that satisfying local constraints does not compromise global layout regularity. The performance gap in OAR suggests that future efforts should focus on stronger global packing and overflow penalties for dense diagrams while preserving current progress in alignment, containment, and code cleanliness.

\begin{figure}[t]
    \centering
    \includegraphics[width=\linewidth]{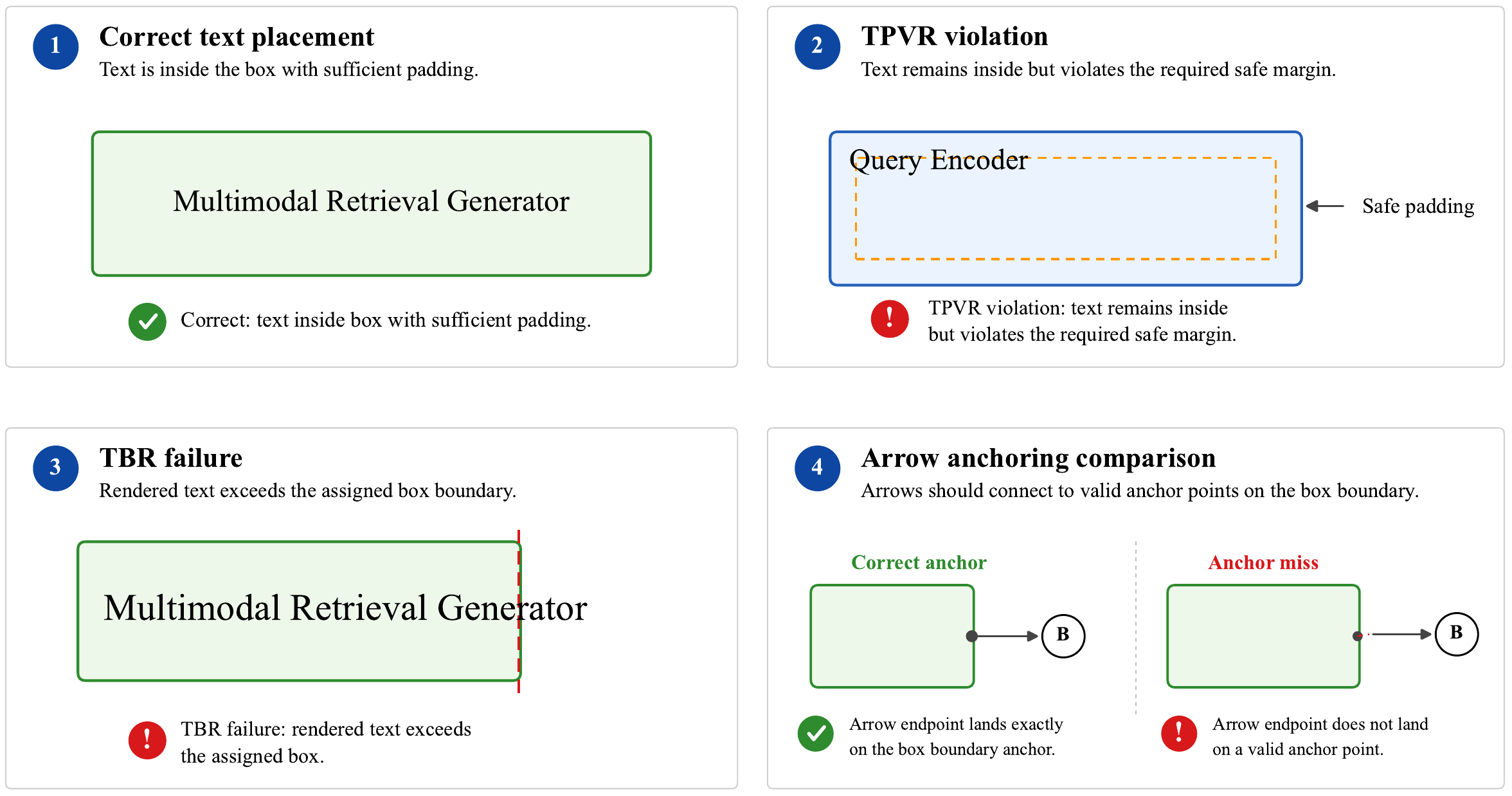}
    \caption{Examples of text containment and arrow anchoring metrics.}
    \vspace{-1.6em}
    \label{fig:tpvr_tbr}
\end{figure}

\subsection{Ablation Study}
\label{sec:ablation}

Table~\ref{tab:ablation_core} presents a controlled ablation study evaluating the GeoSVG-RL framework on a consistent test split using a browser-based verifier and geometry-aware metrics. These experiments isolate four factors: layout-plan conditioning, supervised warm-start, verifier-based reranking, and reinforcement learning (RL) with executable geometric feedback.

Explicitly planning layouts improves all metrics over prompt-only generation by providing coordinate and connectivity priors, though low anchor accuracy and graph consistency show planning alone is insufficient. Supervised fine-tuning (SFT) improves rendering and global stability but fails to resolve local geometric errors. While SFT with planning outperforms the base model on RSR, GFR, and EICR, its AAcc, TBR, and E-F1 scores remain far below those of RL-refined variants, demonstrating that token imitation cannot directly optimize post-rendering constraints for structural usability.

Verifier-based reranking is a competitive intermediate baseline, raising AAcc from 43.2 to 61.7 and E-F1 from 47.3 to 68.4 over SFT. This confirms that the verifier captures quality signals, though inference-time selection underperforms policy refinement which directly optimizes the model distribution. Layout planning and RL are complementary: GeoSVG-RL without planning improves local precision over supervised baselines but underperforms the full model across almost all metrics, particularly GFR and E-F1, proving RL benefits from an explicit geometric contract.

The browser-based verifier is crucial for text-related metrics; an XML-only baseline underperforms on TBR and TPVR because text containment depends on rendered font metrics rather than XML attributes alone. Finally, curriculum weighting and code-cleanliness rewards stabilize RL and improve editability. Furthermore, removing curriculum weighting reduces all metrics, indicating that transitioning from local constraints to global fitting stabilizes training. Similarly, removing cleanliness rewards maintains local geometry but degrades the Clean score, proving geometry rewards alone do not yield well-structured SVG code.

\begin{table}[t]
\centering
\caption{Ablation study on the core components of GeoSVG-RL. All variants are evaluated using the same verifier and test split. Best results are in \textbf{bold}, and second-best results are \underline{underlined}.}
\label{tab:ablation_core}
\resizebox{\linewidth}{!}{
\begin{tabular}{lcccccccccc}
\toprule
\textbf{Model Variant} 
& \textbf{RSR}$\uparrow$ 
& \textbf{GFR}$\uparrow$ 
& \textbf{OAR}$\downarrow$ 
& \textbf{EICR}$\uparrow$ 
& \textbf{AAcc}$\uparrow$ 
& \textbf{AEE}$\downarrow$ 
& \textbf{TBR}$\uparrow$ 
& \textbf{TPVR}$\downarrow$ 
& \textbf{E-F1}$\uparrow$ 
& \textbf{Clean}$\uparrow$ \\
\midrule
Base code model, prompt only 
& 72.4 & 70.8 & 9.8 & 78.5 & 31.7 & 0.245 & 44.6 & 22.8 & 39.2 & 58.4 \\
Base code model + plan prompt 
& 78.6 & 74.5 & 7.4 & 82.1 & 40.5 & 0.210 & 51.8 & 18.6 & 48.5 & 63.1 \\
SFT without layout plan 
& 82.6 & 78.1 & 6.7 & 84.0 & 35.6 & 0.221 & 33.7 & 25.6 & 39.1 & 48.7 \\
SFT with layout plan 
& 88.1 & 83.8 & 5.2 & 88.6 & 43.2 & 0.188 & 39.8 & 19.4 & 47.3 & 50.2 \\
SFT + verifier reranking 
& 90.5 & 87.2 & 3.8 & 91.0 & 61.7 & 0.139 & 63.9 & 9.8 & 68.4 & 70.2 \\
GeoSVG-RL without layout plan 
& 89.0 & 84.1 & 4.9 & 88.8 & 62.4 & 0.142 & 68.5 & 8.7 & 72.9 & 78.0 \\
GeoSVG-RL with XML-only verifier 
& 90.7 & 85.0 & 4.6 & 89.6 & 70.3 & 0.121 & 65.4 & 11.2 & 84.1 & 88.5 \\
GeoSVG-RL without curriculum weighting 
& 91.1 & 86.3 & 4.1 & 90.2 & 74.0 & 0.109 & 77.2 & 5.6 & 87.9 & \cellcolor{green!8}\underline{90.8} \\
GeoSVG-RL without code-cleanliness reward 
& \cellcolor{green!8}\underline{91.6} & \cellcolor{green!8}\underline{89.0} & \cellcolor{green!8}\underline{3.0} & \cellcolor{green!8}\underline{92.8} & \cellcolor{green!8}\underline{78.2} & \cellcolor{green!8}\underline{0.098} & \cellcolor{green!8}\underline{82.6} & \cellcolor{green!8}\underline{3.5} & \cellcolor{green!8}\underline{89.9} & 83.6 \\
\textbf{GeoSVG-RL full} 
& \cellcolor{green!25}\textbf{92.3} & \cellcolor{green!25}\textbf{89.7} & \cellcolor{green!25}\textbf{2.9} & \cellcolor{green!25}\textbf{93.4} & \cellcolor{green!25}\textbf{78.6} & \cellcolor{green!25}\textbf{0.096} & \cellcolor{green!25}\textbf{83.0} & \cellcolor{green!25}\textbf{3.3} & \cellcolor{green!25}\textbf{90.4} & \cellcolor{green!25}\textbf{91.8} \\
\bottomrule
\end{tabular}
}
\vspace{-2em}
\end{table}

\section{Conclusion}
This paper presents a decoupled framework for generating layout-constrained SVG diagrams by incorporating planning, rendering, and verification into the training process through Group Relative Policy Optimization. Our results show that executable geometric feedback successfully bridges the gap between visually plausible outputs and structurally reliable vector programs, leading to significant gains in local precision over the supervised warm start. Although these improvements are notable, the system is still limited by its dependence on synthetic data and a tendency toward global overflow. This suggests that local geometric accuracy currently develops faster than the capacity for global layout optimization. While \method\ improves the controllability of technical illustrations, the risk of producing polished yet inaccurate explanations still requires human oversight. Nevertheless, we believe that verifier-backed constraints provide an essential safeguard by making failure modes measurable and allowing for the systematic auditing of automated design tools.

\bibliographystyle{plain}
\bibliography{refs}


\newpage
\appendix

\section{Implementation Details}
\label{app:implementation_details}

\subsection{Base Model}
The SVG generator is initialized from \texttt{Qwen2.5-Coder-7B-Instruct}, a pretrained autoregressive code model with \texttt{7B} parameters. This model was chosen for its exposure to structured markup languages like XML, HTML, and SVG code during pretraining. Unless otherwise specified, all experiments use LoRA for parameter-efficient fine-tuning. We set the rank to 16, the scaling factor to 32, and the dropout rate to 0.05. Adapters are applied to the \texttt{q\_proj}, \texttt{k\_proj}, \texttt{v\_proj}, \texttt{o\_proj}, \texttt{gate\_proj}, \texttt{up\_proj}, and \texttt{down\_proj} layers.

\subsection{Synthetic Data Generation}
The synthetic corpus consists of 48,000 training, 4,000 validation, and 8,200 test samples. The dataset covers 6 diagram categories: horizontal pipelines, stacked modules, branching flows, grouped containers, retrieval-oriented architectures, and multi-stage ML workflows. Each sample contains a textual prompt, a structured layout plan, a reference SVG program, a rendered PNG, and geometric metadata (node boxes, text regions, connector anchors, and edge lists). The number of nodes ranges from 3 to 8, and directed edges range from 2 to 10. To test generalization, we construct IID, template-held-out, complexity-held-out, and real-prompt test splits, using unseen graph templates, prompt templates, node labels, and layout configurations.

\subsection{Supervised Warm Start}
In the supervised warm-start phase, we train the planner and generator on synthetic prompt-plan-SVG tuples. The planner uses token-level cross-entropy loss over JSON layout plans, while the generator is trained with next-token cross-entropy. Training runs for 3 epochs with a batch size of 128, a learning rate of \texttt{2e-5}, weight decay of 0.01, and a warmup ratio of 0.03, using a maximum sequence length of 4,096 tokens. We select the final checkpoint based on the validation mean geometry score on a consistent held-out split.

\subsection{Browser-backed Verifier}
SVG outputs are rendered using \texttt{Chromium} in headless mode. The rendering canvas is set to 800 $\times$ 600 pixels unless the prompt specifies otherwise. We use \texttt{Arial} with font sizes from 12 to 20 px. Text bounding boxes are retrieved via \texttt{getBBox()} and \texttt{getBoundingClientRect()}, while geometric primitives and transformations are parsed from the SVG XML using \texttt{lxml}. For path-based connectors, start and end points are estimated by sampling the normalized path at its first and last visible coordinates after resolving transformations. A generation is valid only if it successfully renders within a 5-second timeout.

\subsection{GRPO Refinement}
Starting from the warm-start checkpoint, the generator is refined using GRPO. We sample $G=4$ candidates per prompt with a temperature of 0.6, top-$p$ of 0.90, and a limit of 2,048 tokens. Rewards are normalized within each group to compute advantages. We use a clipping coefficient $\eta=0.2$, a learning rate of \texttt{5e-6}, a batch size of 32, and 4 gradient accumulation steps over 1,500 updates. A KL penalty (coefficient 0.02) is applied relative to the reference policy. The final checkpoint is selected by the validation mean geometry score, defined as the average of normalized RSR, GFR, EICR, AAcc, TBR, E-F1, and Clean scores, minus the normalized OAR, AEE, and TPVR.

\subsection{Compute Resources}
Experiments were conducted on NVIDIA A100 GPUs (80GB). The warm-start phase took roughly 96 GPU-hours, and GRPO refinement took 128 GPU-hours. The verifier runs on CPU, processing about 18 SVGs per second using 8 parallel workers.

\subsection{Reward Function}
The total reward is a weighted sum of executable and geometry-aware components:
\[
R(\hat{y}) = \sum_k{\lambda_{k} R_{k}}.
\]
Weights are set as shown in Table~\ref{tab:wts}.
\begin{table}[h]
\centering
\caption{Specific value settings of weights.}
\label{tab:wts}
\setlength{\tabcolsep}{3mm}{
\begin{tabular}{l|cccccccc}
\toprule
\textbf{$\lambda_k$} & $\lambda_{\mathrm{exec}}$ & $\lambda_{\mathrm{fit}}$ & $\lambda_{\mathrm{overflow}}$ & $\lambda_{\mathrm{anchor}}$ & $\lambda_{\mathrm{text}}$ & $\lambda_{\mathrm{padding}}$ & $\lambda_{\mathrm{graph}}$ & $\lambda_{\mathrm{clean}}$ \\
\midrule
\textbf{Value} & 1.00 & 0.60 & 0.50 & 1.20 & 1.10 & 0.50 & 0.90 & 0.30 \\
\bottomrule
\end{tabular}
}
\end{table}

The anchor threshold $\tau$ is 12 px, the required text padding $p$ is 6 px, and the distance normalization factor $d_k$ is the diagonal length of the node box associated with the endpoint $k$. Components are normalized to $[0, 1]$ for rewards or $[-1, 0]$ for penalties. We use a curriculum schedule: initially focusing on anchor, text, padding, and graph terms for the first 500 updates, then linearly increasing the weights for canvas-fit and overflow terms over the next 500 updates.

\subsection{Definition of Metrics}
\begin{table}[h]
    \caption{Geometry-aware evaluation suite for layout-constrained SVG diagram generation.}
    \label{tab:metrics}
    \centering
    \small
    \resizebox{\textwidth}{!}{
    \begin{tabular}{p{2.1cm}p{0.8cm}p{10.9cm}}
    \toprule
    Metric & Better & Principle \\
    \midrule
    RSR & $\uparrow$ & Render Success Rate. Measures whether the SVG parses and renders successfully. \\
    GFR & $\uparrow$ & Global Fit Rate. Checks whether the union of elements lies inside the canvas. \\
    OAR & $\downarrow$ & Overflow Area Ratio. Measures the fraction of area outside the canvas. \\
    EICR & $\uparrow$ & Element-In-Canvas Rate. Percentage of elements fully inside the canvas. \\
    AAcc & $\uparrow$ & Arrow Anchor Accuracy. Percentage of correctly anchored connectors. \\
    AEE & $\downarrow$ & Anchor Endpoint Error. Normalized distance between endpoints and target anchors. \\
    TBR & $\uparrow$ & Text-In-Box Rate. Percentage of text boxes remaining inside containers. \\
    TPVR & $\downarrow$ & Text Padding Violation Rate. Percentage of instances failing the margin threshold. \\
    E-F1 & $\uparrow$ & Edge Connectivity F1. Graph-level F1 score over node connectivity. \\
    Clean & $\uparrow$ & SVG Cleanliness. Rewards semantic primitives and penalizes fragmented code. \\
    \bottomrule
    \end{tabular}
    }
\end{table}

\section{Details of Synthetic Diagram Generation}
Our synthetic data generator samples from various diagram families, including horizontal pipelines, stacked modules, branching structures, and grouped containers. For each instance, the generator selects a graph template, assigns module labels, determines a box-level layout, and produces four aligned artifacts: a textual prompt, a layout plan, a ground-truth SVG, and a rendered PNG. We also generate corrupted variants by intentionally shifting boxes, displacing connector endpoints, shrinking text regions, or extending the layout beyond the canvas. These samples serve as benchmarks for verifying the reliability of the verifier and may support future refinement-based training.

\subsection{Diagram Families}
The current scope includes simple pipelines, dual-branch architectures, grouped blocks, and retrieval-oriented process diagrams. This set is restricted to ensure precise control while remaining broad enough to capture the fundamental box-arrow-text interactions that determine our target metrics.

\subsection{Prompt Construction}
Each synthetic sample is associated with a textual instruction derived from the underlying graph and layout specification. We introduce linguistic variation through template slots, such as order descriptors, stage designations, and relational verbs. This variation is intended to provide sufficient paraphrastic diversity to prevent the model from overfitting to a specific prompt style, rather than for linguistic variety itself.

\section{Verifier Implementation Details}
The verifier combines browser-based rendering with structural parsing. Browser rendering ensures the faithful representation of text layouts, markers, and bounding boxes, which are critical for text-related metrics. Simultaneously, structural parsing extracts editable primitives, transformations, and graph structures from the SVG XML. This dual-path approach is intentional, as certain constraints are best evaluated in the rendered domain, while others are more effectively assessed via the vector program.

\subsection{Text Measurement}
Text-related metrics are computed using rendered bounding boxes rather than raw token counts or simplified width heuristics. This precision is necessary because identical text may occupy different spatial dimensions depending on font size, anchoring, and alignment settings.

\subsection{Connector Recovery}
For connectors based on lines and polylines, endpoints are extracted directly after resolving all transformations. For path-based connectors, we approximate start and end points through normalized path commands. This complexity justifies our decision to reward code cleanliness and prioritize semantic primitives where possible.

\subsection{Graph Extraction}
Recovered nodes and connectors are converted into an induced graph by matching connector endpoints to corresponding node anchor regions. This extracted graph is then used to calculate edge-level precision, recall, and F1 scores.

\section{Reward Design Considerations}
The reward function is modular, with each term corresponding to a specific, independently inspectable failure mode.

\subsection{Limitations of Unitary Visual Scores}
Global image similarity scores cannot reliably distinguish between a connector that terminates on the wrong side of a box and one that aligns precisely with the target anchor. Similarly, such scores are often insensitive to labels that remain within a box but violate safety margins. Our modular reward architecture addresses these limitations by aligning each metric with a concrete structural requirement.

\subsection{Selection of Reward Weights}
Reward terms are first normalized to comparable ranges. We then assign weights based on task priority, with rendering validity, local geometric correctness, global fitting, and code cleanliness in descending order of importance. These coefficients are adjusted using a validation set and confirmed through ablation studies. In practice, we found that increasing the weights for global fitting too early can destabilize training; these adjustments are more effective once local convergence is achieved.

\section{Practical Implementation of GRPO}
Group Relative Policy Optimization (GRPO) is well-suited for this task because the verifier provides an external scalar reward and allows for a direct comparison of multiple SVG candidates for a single prompt. Furthermore, group-relative normalization reduces sensitivity to raw reward scales across different prompts.

\subsection{Rollout Patterns}
For each prompt, we sample a small group of SVG candidates, verify each instance, and compute relative advantages within the group. This is particularly useful when prompt difficulty varies; for instance, a reward of $0.7$ might represent excellence for one prompt but mediocrity for another. Group-relative ranking maintains a localized definition of improvement.

\subsection{Comparison with REINFORCE}
While the REINFORCE algorithm is straightforward, it often exhibits high variance when applied to long SVG sequences with delayed rewards. GRPO provides a stable, clipped update mechanism similar to PPO without requiring a separate critic, offering an efficient approach for verifier-backed SVG generation.

\section{Extended Discussion of Failure Modes}
The most prevalent failure mode remains global overflow. After reinforcement learning, the model shows improved accuracy in arrow placement and text containment, yet some outputs still exhibit aggressive box allocation or drift too close to the canvas boundaries. This suggests that local geometric competence has advanced faster than the capacity for global layout optimization.

A secondary failure mode involves style inconsistency. When the model uses path-based or transformed elements, graph connectivity may remain accurate even if code cleanliness declines. This confirms the need for a code-level cleanliness term in the reward function, despite its lower weight relative to primary geometric terms.

\section{Prompt and Plan Schema}
The planner output is simplified for clarity. Each sample includes a canvas specification, a list of node objects, and a list of directed connectors. A node object records its type, bounding box, and textual label, while a connector specifies the source node, target node, and anchor types. This representation is expressive enough for box-arrow-text diagrams while remaining suitable for automatic verification.

Prompts describe high-level structures rather than pixel-level coordinates, specifying stage sequences, branching, and semantic roles. The planner is responsible for translating these semantic instructions into explicit coordinates.

\section{Hyperparameters}
\label{app:hyperparameters}

Table~\ref{tab:hyperparameters} summarizes the main training and evaluation hyperparameters used in GeoSVG-RL. Unless otherwise specified, the same hyperparameters are used for all ablation variants to ensure that differences in performance reflect architectural or objective-level changes rather than tuning differences.

\begin{table}[t]
\centering
\caption{Training and evaluation hyperparameters used in GeoSVG-RL.}
\label{tab:hyperparameters}
\setlength{\tabcolsep}{10mm}{
\begin{tabular}{ll}
\toprule
\textbf{Hyperparameter} & \textbf{Value} \\
\midrule
Base model & Qwen2.5-Coder-7B-Instruct \\
Number of parameters & 7B \\
Fine-tuning method & LoRA \\
LoRA rank & 16 \\
LoRA alpha & 32 \\
LoRA dropout & 0.05 \\
Maximum sequence length & 4096 \\
SFT epochs & 3 \\
SFT batch size & 128 \\
SFT learning rate &\texttt{2e-5} \\
SFT warmup ratio & 0.03 \\
SFT weight decay & 0.01 \\
GRPO group size $G$ & 4 \\
GRPO clipping range $\eta$ & 0.2 \\
GRPO learning rate & \texttt{5e-6} \\
GRPO batch size & 32 \\
GRPO updates & 1500 \\
Sampling temperature & 0.6 \\
Top-$p$ & 0.90 \\
Maximum generation tokens & 2048 \\
KL coefficient & 0.02 \\
Anchor threshold & 12 px \\
Text padding threshold $p$ & 6 px \\
Canvas size & 800 $\times$ 600 \\
Browser engine & Chromium 124.0 \\
SVG parser & \texttt{lxml} \\
Number of browser workers & 8 \\
Random seeds & 13, 21, 42 \\
\bottomrule
\end{tabular}
}
\end{table}

\section{Dataset Statistics}
\label{app:dataset_statistics}

Table~\ref{tab:dataset_statistics} reports the number of prompts, layout plans, and SVG programs in each split. The IID test split follows the same generator distribution as the training set but uses disjoint random seeds and prompt variations. The template-held-out split evaluates unseen graph templates and layout patterns. The complexity-held-out split contains diagrams with more nodes, edges, branches, and grouped components than those typically observed during training. The real-prompt split contains manually collected prompts inspired by common technical diagrams in machine learning papers, system architecture figures, and workflow illustrations.

\begin{table}[t]
\centering
\caption{Statistics of the synthetic diagram corpus.}
\label{tab:dataset_statistics}
\begin{tabular}{lccc}
\toprule
\textbf{Split} 
& \textbf{\# Prompts} 
& \textbf{\# Layout Plans} 
& \textbf{\# SVG Programs} \\
\midrule
Train & 48,000 & 48,000 & 48,000 \\
Validation & 4,000 & 4,000 & 4,000 \\
IID test & 4,000 & 4,000 & 4,000 \\
Template-held-out test & 2,000 & 2,000 & 2,000 \\
Complexity-held-out test & 2,000 & 2,000 & 2,000 \\
Real-prompt test & 200 & 200 & 200 \\
\bottomrule
\end{tabular}
\end{table}

Table~\ref{tab:diagram_complexity} summarizes the structural complexity of each split. The complexity-held-out and real-prompt splits are intentionally more difficult than the IID test split, providing a stress test for global layout control and graph consistency.

\begin{table}[t]
\centering
\caption{Diagram complexity statistics. Values are reported as ranges, with mean values in parentheses when applicable.}
\label{tab:diagram_complexity}
\resizebox{\linewidth}{!}{
\begin{tabular}{lcccccc}
\toprule
\textbf{Split} 
& \textbf{Nodes} 
& \textbf{Edges} 
& \textbf{Text boxes} 
& \textbf{Branches} 
& \textbf{Groups} 
& \textbf{Canvas sizes} \\
\midrule
Train & 3--7 (5.1) & 2--8 (4.6) & 3--10 (6.2) & 0--2 (0.7) & 0--2 (0.6) & 600$\times$400, 800$\times$600 \\
Validation & 3--7 (5.0) & 2--8 (4.5) & 3--10 (6.1) & 0--2 (0.7) & 0--2 (0.6) & 600$\times$400, 800$\times$600 \\
IID test & 3--7 (5.1) & 2--8 (4.6) & 3--10 (6.2) & 0--2 (0.7) & 0--2 (0.6) & 600$\times$400, 800$\times$600 \\
Template-held-out test & 3--7 (5.3) & 2--9 (4.9) & 3--11 (6.5) & 0--3 (0.9) & 0--2 (0.7) & 600$\times$400, 800$\times$600 \\
Complexity-held-out test & 6--10 (7.8) & 6--13 (9.1) & 6--14 (9.4) & 1--4 (2.1) & 1--3 (1.4) & 800$\times$600, 1000$\times$700 \\
Real-prompt test & 4--9 (6.4) & 3--12 (6.8) & 4--13 (7.9) & 0--4 (1.6) & 0--3 (1.1) & 800$\times$600, 1000$\times$700 \\
\bottomrule
\end{tabular}
}
\end{table}

\section{Additional Ablation Studies}
\label{app:additional_ablation}

\subsection{Reward Component Ablation}
\label{app:reward_ablation}

Table~\ref{tab:ablation_reward} studies the contribution of each reward component by removing one term at a time from the full GeoSVG-RL objective. The model, training data, verifier, and GRPO hyperparameters are kept fixed across all variants. This controlled setting allows us to identify which failure modes are addressed by each reward term.

\begin{table}[t]
\centering
\caption{Reward component ablation. Each variant removes one reward component from the full GeoSVG-RL objective while keeping the same model, data, verifier, and GRPO hyperparameters. Best results are in \textbf{bold}, and second-best results are \underline{underlined}.}
\label{tab:ablation_reward}
\resizebox{\linewidth}{!}{
\begin{tabular}{lcccccccccc}
\toprule
\textbf{Reward Variant} 
& \textbf{RSR}$\uparrow$ 
& \textbf{GFR}$\uparrow$ 
& \textbf{OAR}$\downarrow$ 
& \textbf{EICR}$\uparrow$ 
& \textbf{AAcc}$\uparrow$ 
& \textbf{AEE}$\downarrow$ 
& \textbf{TBR}$\uparrow$ 
& \textbf{TPVR}$\downarrow$ 
& \textbf{E-F1}$\uparrow$ 
& \textbf{Clean}$\uparrow$ \\
\midrule
\textbf{GeoSVG-RL full} 
& \cellcolor{green!25}\textbf{92.3} & \cellcolor{green!25}\textbf{89.7} & \cellcolor{green!25}\textbf{2.9} & \cellcolor{green!25}\textbf{93.4} & \cellcolor{green!25}\textbf{78.6} & \cellcolor{green!25}\textbf{0.096} & \cellcolor{green!25}\textbf{83.0} & \cellcolor{green!25}\textbf{3.3} & \cellcolor{green!25}\textbf{90.4} & \cellcolor{green!25}\textbf{91.8} \\
w/o rendering-validity reward $R_{\mathrm{exec}}$ 
& 84.9 & 82.4 & 5.6 & 87.1 & 74.8 & 0.105 & 79.6 & 4.8 & 87.6 & 89.9 \\
w/o canvas-fit reward $R_{\mathrm{fit}}$ 
& 91.7 & 82.8 & 6.7 & 87.9 & 78.1 & 0.098 & 82.4 & 3.7 & 89.9 & 91.0 \\
w/o overflow penalty $R_{\mathrm{overflow}}$ 
& 92.0 & 84.6 & 6.1 & 88.8 & 78.0 & 0.099 & 82.1 & 3.9 & 89.7 & 91.2 \\
w/o anchor reward $R_{\mathrm{anchor}}$ 
& 91.8 & 88.9 & 3.2 & 92.7 & 61.5 & 0.162 & 82.0 & 3.8 & 86.8 & 91.4 \\
w/o text-containment reward $R_{\mathrm{text}}$ 
& 91.9 & \cellcolor{green!8}\underline{89.1} & 3.1 & 92.9 & 77.9 & 0.098 & 62.7 & 12.4 & 89.3 & 91.1 \\
w/o padding reward $R_{\mathrm{padding}}$ 
& \cellcolor{green!8}\underline{92.1} & 89.0 & \cellcolor{green!8}\underline{3.0} & \cellcolor{green!8}\underline{93.0} & \cellcolor{green!8}\underline{78.2} & \cellcolor{green!8}\underline{0.097} & 80.4 & 9.6 & \cellcolor{green!8}\underline{90.0} & \cellcolor{green!8}\underline{91.5} \\
w/o graph reward $R_{\mathrm{graph}}$ 
& 91.6 & 88.8 & 3.3 & 92.6 & 76.9 & 0.101 & 81.7 & 3.9 & 72.8 & 90.8 \\
w/o cleanliness reward $R_{\mathrm{clean}}$ 
& 91.6 & 89.0 & \cellcolor{green!8}\underline{3.0} & 92.8 & \cellcolor{green!8}\underline{78.2} & 0.098 & \cellcolor{green!8}\underline{82.6} & \cellcolor{green!8}\underline{3.5} & 89.9 & 83.6 \\
w/o reward normalization 
& 89.8 & 85.7 & 4.4 & 90.1 & 70.8 & 0.123 & 72.6 & 8.1 & 82.3 & 87.0 \\
\bottomrule
\end{tabular}
}
\end{table}

Removing the rendering-validity reward causes the largest drop in RSR, confirming that malformed or partially renderable SVG programs remain a significant source of failure during RL. Removing the canvas-fit reward or overflow penalty mainly affects GFR, OAR, and EICR, while leaving local metrics such as AAcc and TBR relatively stable. This indicates that local geometric correctness and global canvas packing are related but separable failure modes.

The anchor and text rewards show strong metric-specific effects. Without the anchor reward, AAcc decreases from 78.6 to 61.5 and AEE increases from 0.096 to 0.162. Without the text-containment reward, TBR decreases from 83.0 to 62.7 and TPVR increases from 3.3 to 12.4. These changes confirm that the corresponding reward terms are necessary for optimizing the local constraints that are most difficult to capture through token-level training alone.

The graph reward is primarily responsible for preserving diagram connectivity. Removing it reduces E-F1 from 90.4 to 72.8, while other geometric metrics remain comparatively less affected. Similarly, removing the cleanliness reward substantially reduces Clean from 91.8 to 83.6 without severely damaging local geometry. This supports our design of treating code cleanliness as a separate objective: semantically meaningful SVG primitives are not guaranteed by geometric correctness alone.

Finally, removing reward normalization degrades nearly all metrics. This suggests that raw verifier scores have different scales across prompts and reward components. Group-relative optimization benefits from normalized rewards because it makes candidate comparisons more stable across diagrams with different levels of complexity.

\subsection{Optimization and Sampling Ablation}
\label{app:optimization_ablation}

Table~\ref{tab:ablation_optimization} compares different policy optimization strategies and sampling configurations. All variants start from the same supervised warm-start checkpoint and use the same verifier. This experiment is intended to isolate the effect of the RL optimizer and the group-based sampling strategy.

\begin{table}[t]
\centering
\caption{Optimization and sampling ablation. We compare different policy optimization strategies and group sizes while keeping the same supervised warm-start checkpoint and verifier.
}
\label{tab:ablation_optimization}
\resizebox{\linewidth}{!}{
\begin{tabular}{lcccccccccc}
\toprule
\textbf{Optimization Variant} 
& \textbf{RSR}$\uparrow$ 
& \textbf{GFR}$\uparrow$ 
& \textbf{OAR}$\downarrow$ 
& \textbf{EICR}$\uparrow$ 
& \textbf{AAcc}$\uparrow$ 
& \textbf{AEE}$\downarrow$ 
& \textbf{TBR}$\uparrow$ 
& \textbf{TPVR}$\downarrow$ 
& \textbf{E-F1}$\uparrow$ 
& \textbf{Clean}$\uparrow$ \\
\midrule
Supervised warm start 
& 88.1 & 83.8 & 5.2 & 88.6 & 43.2 & 0.188 & 39.8 & 19.4 & 47.3 & 50.2 \\
REINFORCE 
& 88.7 & 84.0 & 5.5 & 88.3 & 58.9 & 0.151 & 61.2 & 11.8 & 68.0 & 71.4 \\
PPO-style update with value head 
& 90.8 & 86.4 & 4.0 & 91.0 & 72.5 & 0.113 & 75.8 & 6.4 & 84.6 & 88.9 \\
GRPO, group size $G=2$ 
& 91.2 & 87.6 & 3.6 & 91.8 & 74.4 & 0.108 & 77.9 & 5.5 & 86.7 & 90.0 \\
GRPO, group size $G=4$ 
& \cellcolor{green!25}\textbf{92.3} & \cellcolor{green!25}\textbf{89.7} & \cellcolor{green!25}\textbf{2.9} & \cellcolor{green!25}\textbf{93.4} & \cellcolor{green!25}\textbf{78.6} & \cellcolor{green!25}\textbf{0.096} & \cellcolor{green!25}\textbf{83.0} & \cellcolor{green!25}\textbf{3.3} & \cellcolor{green!25}\textbf{90.4} & \cellcolor{green!25}\textbf{91.8} \\
GRPO, group size $G=8$ 
& 91.4 & 87.9 & 3.8 & 92.1 & 75.9 & 0.107 & 78.5 & 5.1 & 87.8 & 90.4 \\
GRPO, high-temperature sampling 
& 89.6 & 84.7 & 5.1 & 89.0 & 71.0 & 0.121 & 72.4 & 7.9 & 82.5 & 86.7 \\
GRPO, low-temperature sampling 
& \cellcolor{green!8}\underline{92.2} & \cellcolor{green!8}\underline{89.5} & \cellcolor{green!8}\underline{3.0} & \cellcolor{green!8}\underline{93.3} & \cellcolor{green!8}\underline{78.4} & \cellcolor{green!8}\underline{0.097} & \cellcolor{green!8}\underline{82.8} & \cellcolor{green!8}\underline{3.4} & \cellcolor{green!8}\underline{90.2} & \cellcolor{green!8}\underline{91.6} \\
\bottomrule
\end{tabular}
}
\end{table}

REINFORCE improves over the supervised warm start on local geometric metrics, but the gains are limited and global layout metrics remain unstable. This is consistent with the high-variance nature of sequence-level rewards for long SVG programs. PPO-style optimization with a value head performs better, especially on AAcc, TBR, and E-F1, but it requires training an additional critic and remains below GRPO.

GRPO provides the best trade-off between stability and implementation simplicity. Group-relative normalization makes the reward comparison local to each prompt, reducing sensitivity to differences in diagram difficulty. A group size of $G=4$ performs best in our setting. Using $G=2$ provides weaker candidate contrast, while $G=8$ increases sampling diversity but introduces more unstable low-quality candidates during training. High-temperature sampling similarly degrades performance because it increases syntactic and geometric failure rates. Low-temperature sampling is close to the full configuration, but the full setting with moderate temperature and group size $G=4$ achieves the best overall performance.

\subsection{Reproducibility}
\label{app:reproducibility}

All experiments use fixed random seeds 13, 21, and 42. The reported main results use the checkpoint selected based on the validation mean geometry score. For evaluation, each model generates one SVG program per prompt without self-consistency, majority voting, or manual post-processing. Non-SVG explanatory text is removed before verification. Failed generations are assigned zero for geometry-dependent success metrics, while distance- and overflow-based metrics are computed only when rendering succeeds. The same browser-backed verifier, parser, canvas specification, and metric implementation are used for all methods and ablation variants.



\end{document}